\documentclass{article}
\usepackage[utf8]{inputenc}
\usepackage{graphicx}
\usepackage{multirow}
\usepackage{url}
\usepackage{hyperref}
\usepackage{xcolor}
\usepackage{arxiv}

\usepackage{hyperref}
\hypersetup{
pdftitle={Experimental Energy Consumption Analysis of a Flapping-Wing Robot},
pdfauthor={R. Tapia, A.C. Satue, S.R. Nekoo, J. Martínez-de Dios, and A. Ollero},
pdfsubject={2023 IEEE International Conference on Robotics and Automation. Workshop on Energy Efficient Aerial Robotic Systems}
}

\title{Experimental Energy Consumption Analysis of a Flapping-Wing Robot}

\author{
\href{https://orcid.org/0000-0002-4435-5466}{\includegraphics[scale=0.06]{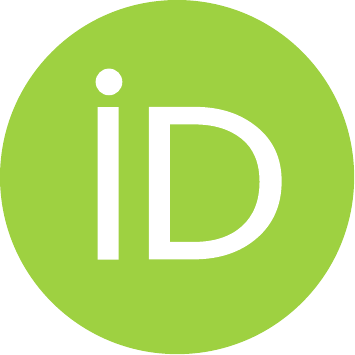}\hspace{1mm}Raul Tapia}\\
	GRVC Robotics Lab.\\
	Universidad de Sevilla\\
	\hphantom{00}\texttt{raultapia@us.es}\hphantom{00}\\
	\And
Alvaro Cesar Satue \\
	GRVC Robotics Lab.\\
	Universidad de Sevilla\\
	\hphantom{00}\texttt{alvsatcre@alum.us.es}\hphantom{00}\\
	\And
\href{https://orcid.org/0000-0003-1396-5082}{\includegraphics[scale=0.06]{orcid.pdf}\hspace{1mm}Saeed Rafee Nekoo} \\
	GRVC Robotics Lab.\\
	Universidad de Sevilla\\
	\hphantom{00}\texttt{saerafee@yahoo.com}\hphantom{00}\\
	\And
\href{https://orcid.org/0000-0001-9431-7831}{\includegraphics[scale=0.06]{orcid.pdf}\hspace{1mm}José Ramiro Martínez-de Dios} \\
	GRVC Robotics Lab.\\
	Universidad de Sevilla\\
	\texttt{jdedios@us.es}\\
	\And
\href{https://orcid.org/0000-0003-2155-2472}{\includegraphics[scale=0.06]{orcid.pdf}\hspace{1mm}Anibal Ollero} \\
	GRVC Robotics Lab.\\
	Universidad de Sevilla\\
	\texttt{aollero@us.es}\\
}

\paperurl{https://raultapia.com/preprints/0003/redirect}
\journal{International Conference on Robotics and Automation 2023 Workshop on Energy Efficient Aerial Robotic Systems}
\editorial{IEEE}

\begin{document}

\maketitle

\begin{abstract}
One of the motivations for exploring flapping-wing aerial robotic systems is to seek energy reduction,  by maintaining manoeuvrability,  compared to conventional unmanned aerial systems. A Flapping Wing Flying Robot (FWFR) can glide in favourable wind conditions, decreasing energy consumption significantly. In addition, it is also necessary to investigate the power consumption of the components in the flapping-wing robot. In this work, two sets of the FWFR components are analyzed in terms of power consumption: a) motor/electronics components and b) a vision system for monitoring the environment during the flight. A measurement device is used to record the power utilization of the motors in the \textit{launching} and \textit{ascending} phases of the flight and also in \textit{cruising} flight around the desired height. Additionally, an analysis of event cameras and stereo vision systems in terms of energy consumption has been performed. The results provide a first step towards decreasing battery usage and, consequently, providing additional flight time.
\end{abstract}

\section{Introduction}
\label{sec:introduction}
Flapping-wing flying robots (FWFR) are bioinspired platforms that use flapping wings to generate lift and thrust. They have high maneuverability without fast rotating propellers, which makes them less dangerous and more robust against collisions than multirotors. Their potentialities in a wide range of applications have motivated significant R\&D interest in the last years \cite{ollero2022past}. The design of the flapping-wing robots needs a reduction of weight in all possible components of the robot to save payload for other equipment, such as  \cite{nekoo202279}, vision camera for monitoring \cite{rodriguez2022free,pan2020development},  bio-inspired claws \cite{gomez2020sma}, or even manipulators. The current E-Flap robot weighs 500g with almost 500g payload capacity \cite{zufferey2021design}. One important element (relatively heavy $\approx 13.4\%$ of the weight) of the robot is a single battery that provides all the power of the components from the motors to the cameras. Then,  all of the components should  be optimized in terms of power consumption.

Online onboard perception of flapping-wing robots involves strong challenges. In fact, FWFR control and guidance methods have traditionally relied on external sensors such as motion capture systems \cite{ma2013controlled,maldonado2020adaptive}. In the last years, the advances in miniaturization have facilitated the integration of online perception systems on board flapping-wing robots. One of the first was the obstacle avoidance method proposed in \cite{tijmons2017obstacle}, which used a lightweight stereo camera to detect static obstacles using disparity maps. Also, an event-based dynamic obstacle avoidance method for flapping-wing robots was presented in \cite{rodriguez2022free}. An FWFR guidance method based on event cameras was investigated using line tracking algorithms, and a visual-servoing controller \cite{gomez2021why}. However, onboard online perception systems for the autonomous navigation of flapping-wing robots is still an under-researched topic. The paper  \cite{rodriguez2021griffin} presents a  flapping-wing perception dataset,  including measurements recorded from an event camera and a traditional camera. 

\begin{figure}[t]
\centering
\includegraphics[width=0.4\linewidth]{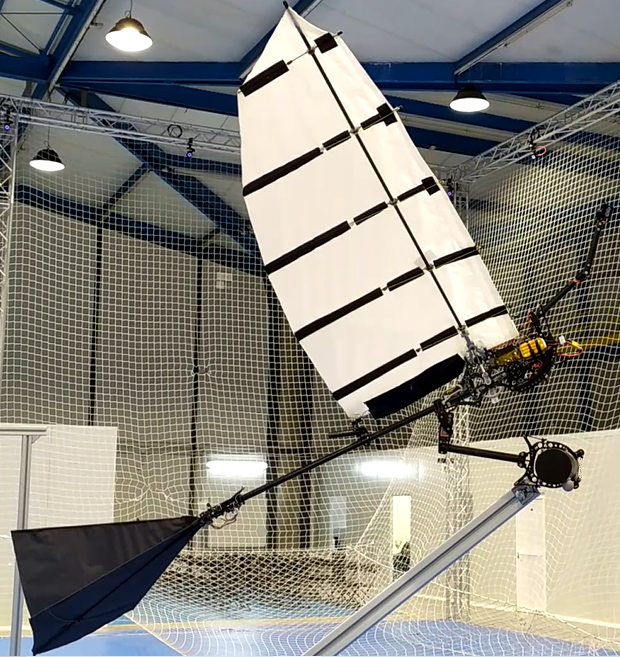}
\caption{The E-Flap robot developed at the GRVC Robotics Lab. of the University of Seville, placed on an artificial branch for manipulation.}
\label{fig:intro}
\end{figure}

Another relevant advantage of FWFR over multirotor drones is their energy efficiency. Flapping-wing platforms can frequently exchange flight phase between flapping-wing and gliding, and this drastically reduces the consumed energy \cite{croon2020flapping}. Energy consumption is critical and depends on the type of flight and also on the onboard sensors and electronics. This has implications not only in flight endurance but also --due to the need for lighter batteries-- in payload and weight distribution. We are interested in developing methods that enable us to predict the consumed energy by autonomous flapping-wing robots in different types of flights using different onboard sensors.

This paper presents an experimental energy consumption analysis of an autonomous flapping-wing robot as a way of predicting the consumed energy in a given type of flight with pre-installed onboard sensors. The analysis includes the energy consumed both for flying and by the onboard perception system. The study includes experiments in test benches and also flight tests conducted with the E-Flap robot developed by the GRVC Robotics Lab. at the University of Sevilla, shown in Figure \ref{fig:intro}.

This paper is structured as follows. The measurement method and description of the  consumption of energy during the FWFR flight are presented in Section \ref{sec:FWFR}. Section \ref{sec:vision} analyses the consumed energy of the onboard perception system. Section \ref{sec:conclusion} closes the paper with the conclusions.

\section{Flapping-Wing Flight Energy Consumption}
\label{sec:FWFR}
\begin{figure}[t]
\centering
\includegraphics[width=0.5\linewidth]{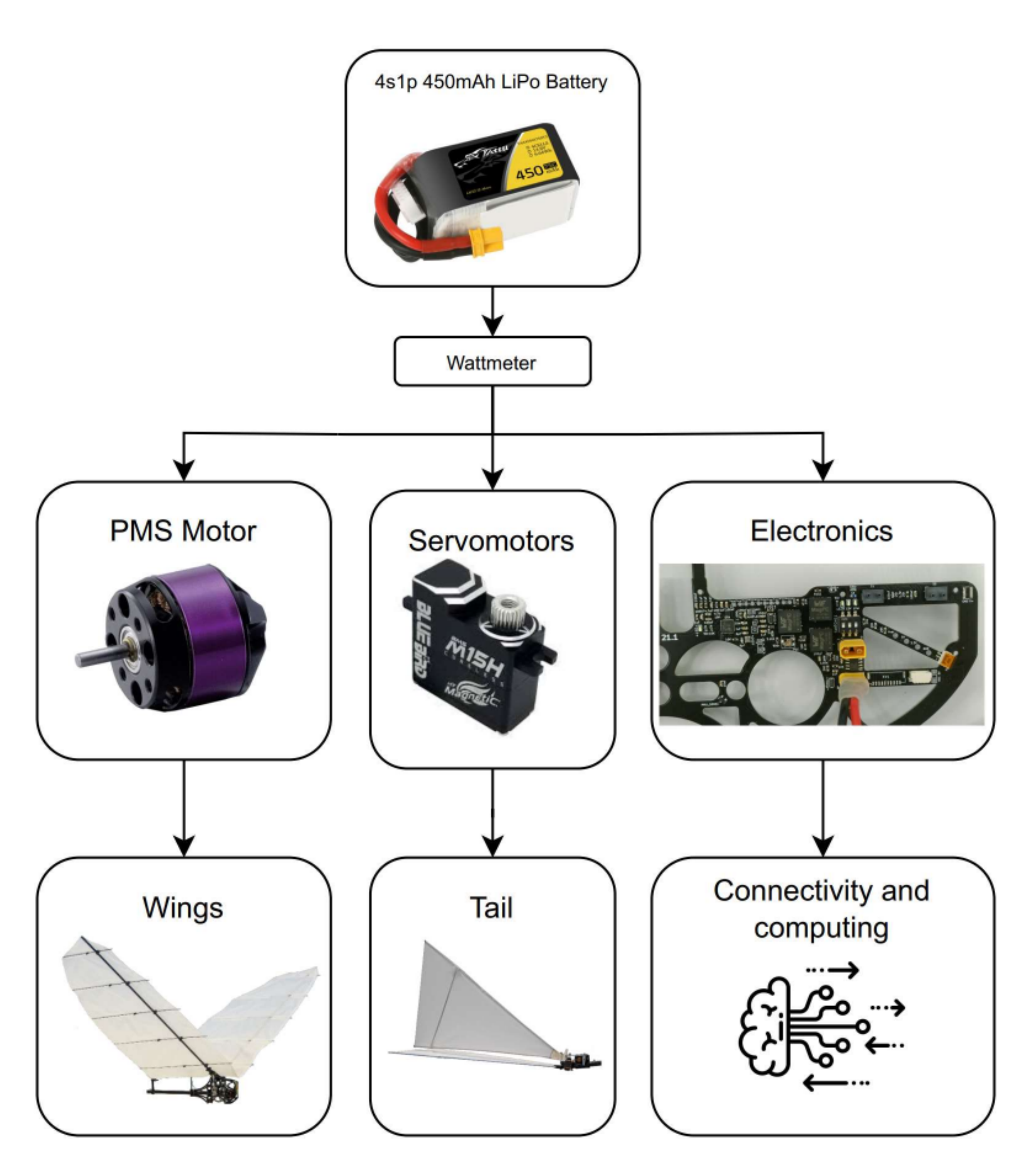}
\caption{The diagram of relevant consumers and source of energy in E-Flap aerial robot prototype.}
\vspace{-0.5em}
\label{fig:Energy_consumers}
\end{figure}

In order to measure the flying power consumption of a bio-inspired aerial robot, it is important to define different components of the platform in the analysis. A detailed description of the platform could be found in \cite{zufferey2021design}. Here we briefly revisit the work with more focus on power management. The source and consumers of energy in the E-Elap robot are shown in Fig. \ref{fig:Energy_consumers}. It includes a four-cell-in-series Lithium Polymer (LiPo) battery. The main consumers are: a flapping permanent magnet synchronous motor (PMSM), tail servomotors for rudder and elevator, and the electronics, mainly composed of regulators, sensors, main computer with WiFi connectivity (Cortex-A7 CPU), and a digital signal processor (DSP) board (Cortex-M4 CPU).

The power consumption of each of the above elements depends mainly on wind  conditions, flight trajectory, and speed. In the indoor test bed there is no external wind as a disturbance; nevertheless, the counteracting flow of the air on the tail and surface of the wings (drag force) affects the consumption of the power in the corresponding actuators. Ambient temperature also augments power consumption during flights. While temperature affects negatively the overall efficiency of electronics, motors, and drivers, wind speed can be favorable or unfavorable with respect to power management depending on the flight trajectory. The conducted experiments were performed in the GRVC Robotics Lab. indoor test bed; hence, the influence of external wind disturbance was not considered in this analysis.

\begin{figure*}[t]
\centering
\includegraphics[width=0.8\linewidth]{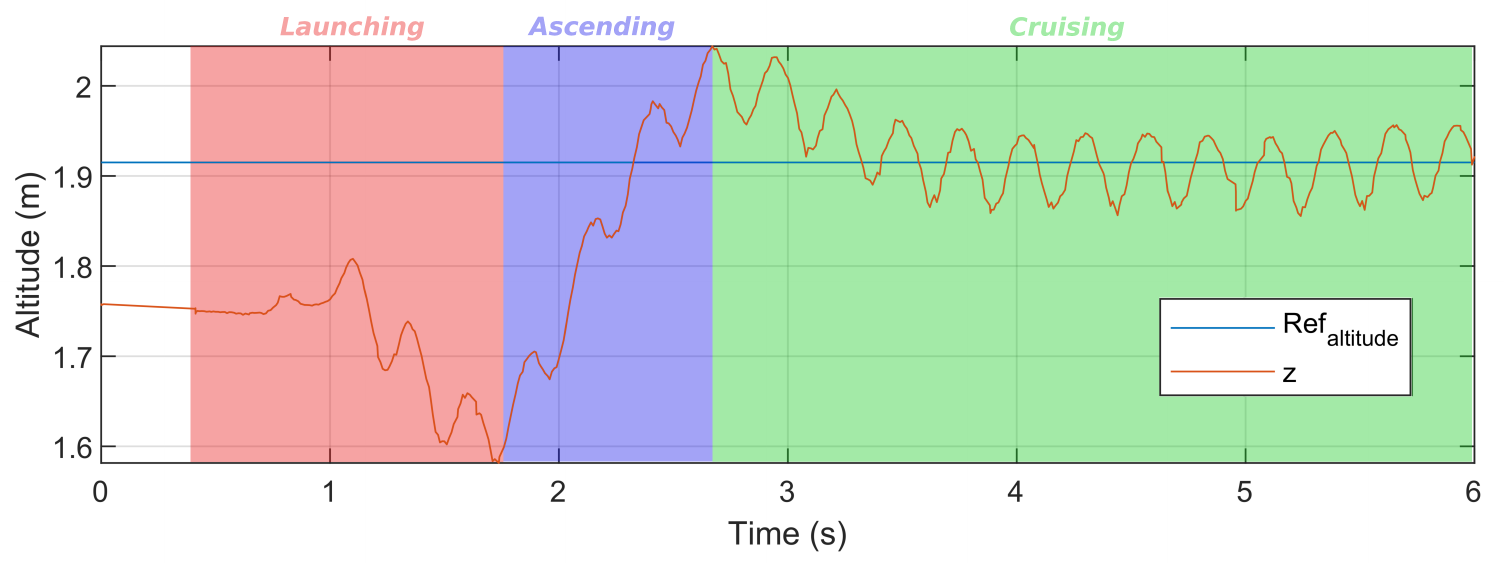}
\vspace{-1em}
\caption{The typical flying profile: it starts with \textit{launching}, continues with \textit{ascending} flight, and finishes with \textit{cruising} flight.}
\label{fig:Fly_stages}
\end{figure*}

The flight trajectory of an aerial robot has a great dependence on energy consumption \cite{moreno2022design}. Two different types of trajectories have been analyzed: \textit{ascending} (i.e. increasing altitude) and \textit{cruising} (keeping altitude around steady-state condition). A set of experiments has been performed and one of the results is shown in Fig. \ref{fig:Fly_stages}. Notice that flapping-wing robots can perform \textit{descending} flights in gliding mode with negligible energy consumption as they could be stabilized without flapping and through the tail, as an actual bird would do. In this analysis, the energy consumption is  measured assuming an approximate constant flapping rate (with changes less than 5\%). The goal of this work is to provide a first approach to flying consumption for our platform at a constant flapping frequency, proposing for further research a more extensive comparison of different trajectories and control strategies.

The consumption of the onboard computer and electronics (without the camera, which will be analyzed separately in Section \ref{sec:vision}) was measured using an INA219-based Gravity I2C digital wattmeter, resulting in less than 2W. The required power from the tail servomotors working around the steady-state position in \textit{cruising} flight is around 1W in average. However, the peaks depending on counteracting airflow and aggressive maneuvers could increase up to 10W. The flapping motor is the main actuator of the altitude control, resulting in the highest power consumption, taking an average of 160W for a \textit{cruising} flight. The recorded data revealed that the flapping wing, servomotors, and rest of the electronics consume 93.02\%, 5.81\%, and 1.16\%, respectively. Even when the tail is working in aggressive maneuvers, the main consumer is the flapping motor, which limits the flying time in \textit{ascending} and \textit{cruising}. A control strategy that optimizes flying range by combining flapping and gliding is under development for further energy analysis. To compute the consumed energy of the flight in each zone of Fig. \ref{fig:Fly_stages}, the following aspects should be regarded. The FWFR starts the flight at an initial speed of 4m/s, provided by an external launcher. This energy of the launching system is not visible in the measured graph of the results. The second aspect is that the flapping exists throughout the whole flight in this preliminary experiment for the indoor test. Since the flight zone is limited to 15m diagonal space in the testbed, the possibility of hybrid flight (gliding and flapping) has not been tried in this work. During the \textit{ascending} flight, the tail servomotors, particularly the elevator, consume a lot of its current, close to saturation to keep the elevator steady against the airflow. This situation does not exist in the cruise flight when the bird is regulated slightly around the desired point. Regarding the mentioned points, and omitting the time of launching from the graph, the flight time is found 5.5s. The consumed energy of the \textit{launching} phase and \textit{ascending} part is found 52.9\% and the cruise flight 47.1\%.

\begin{figure}[t]
\centering
\includegraphics[width=0.8\linewidth]{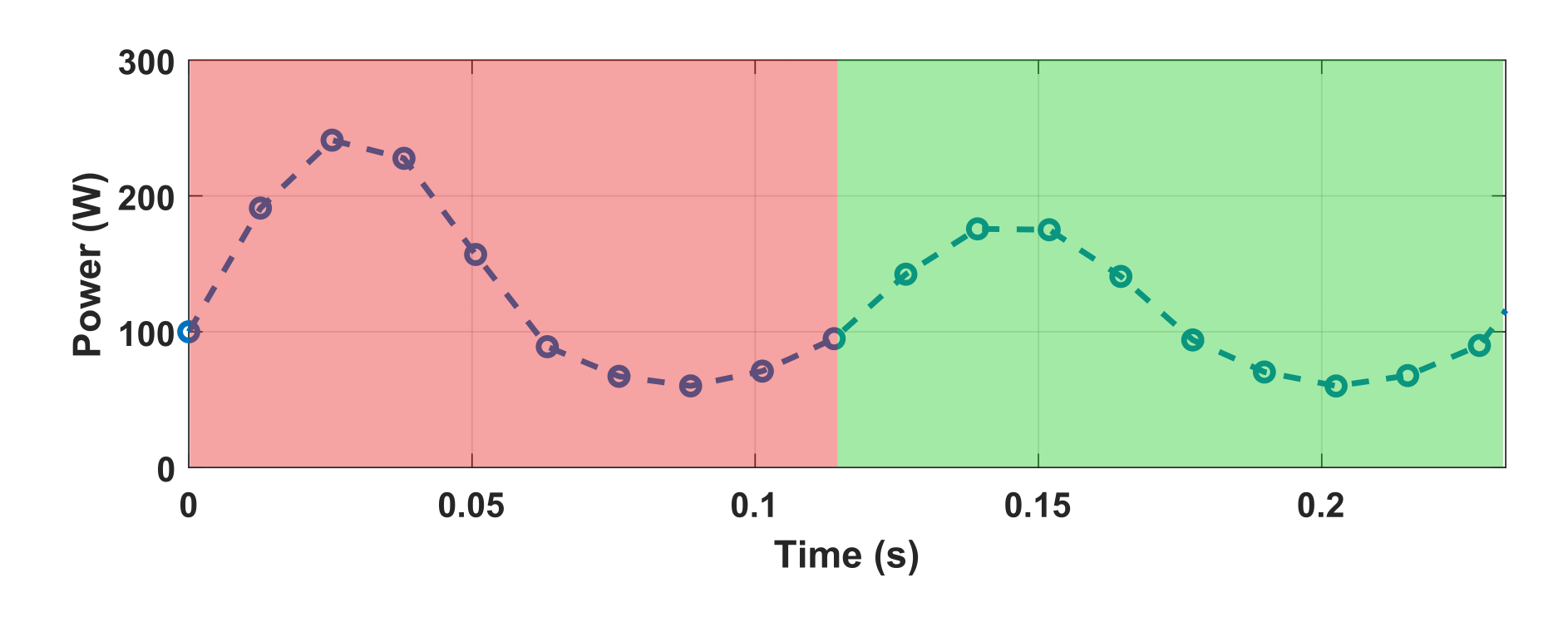}
\vspace{-1.5em}
\caption{Mean power cycle of a flap during the flight shown in Fig. \ref{fig:Fly_stages}: down-stroke (red) and upstroke (green).}
\label{fig:Power-cycle}
\end{figure}

An effect that has been confirmed experimentally corresponds to the positive mean lift force during a flapping cycle. As shown in Fig. \ref{fig:Power-cycle}, maximum consumption is achieved during the down-stroke, while a lower peak consumption is measured during the upstroke. That results in an irregular cyclic discharge of the battery, whose effect on battery lifespan and usable capacity for LiPo chemistry needs to be analyzed in further work.

\section{Perception System Energy Consumption}
\label{sec:vision}
Although the energy consumption of the perception system is much smaller than that of the power components, in some phases of the flight, such as gliding for surveillance applications, it may be relevant. The design of perception systems for FWFRs has strong constraints. First, flapping strokes generate mechanical vibrations and abrupt pitch changes, that can perturb the onboard sensor measurements \cite{gomez2019towards}. Besides, they can glide and flap at different flapping frequencies. Also, flapping-wing robots have strict payload and weight distribution restrictions, which impact the weight and shape of the sensors, electronics, batteries, and other onboard components \cite{zufferey2022ornithopters}. These constraints prevent the use of sensors that are common in multirotor drones, such as LiDARS, which their weight usually exceeds the FWFR's available payload.

Vision sensors have low size and weight. They are suitable for online onboard flapping-wing robot perception, and in fact, all existing related works selected different types of cameras as the main sensor. Flapping-wing robots have employed stereo vision as they mimic the perception of large birds \cite{tijmons2017obstacle}. In addition, stereo vision enables the use of standard perception algorithms and provides depth estimation, which is crucial for applications such as navigation and obstacle avoidance.

Event cameras have gained more and more attention in recent years. These bioinspired sensors provide high responsiveness, dynamic range, and robustness against motion blur and offer significantly lower power consumption compared to standard cameras  \cite{gallego2020event}. This is particularly interesting for flapping-wing robots, where energy efficiency is critical. Although event-based vision requires specialized algorithms that differ from those used in standard vision, there have been significant advancements in event-based vision in recent years, offering  better solutions for visual perception for flapping-wing robots \cite{gomez2021why,rodriguez2022free}.

In this section, we present an experimental analysis of the energy consumption of three widely-used stereo vision systems and one event camera. The cameras considered are 1) a \textit{RealSense D435} stereo-pair widely used in many robotics applications, 2) a large stereo pair with high resolution \textit{Stereolabs ZED}, 3) a small lightweight stereo camera \textit{eCapture G53}, and 4) a \textit{DAVIS346} event camera which contains a Dynamic Vision Sensor (\textit{DAVIS346 DVS}). The camera specifications are presented in Table \ref{tab:cameras}.

\begin{table}[t]
\centering
\caption{Specifications of the evaluated cameras.}
\begin{tabular}{ |c|c|c|c| } 
\hline
\textbf{Name} & \textbf{FPS} & \textbf{Resolution} & \textbf{Ideal Depth Range} \\
\hline
\textit{RealSense D435} & 30 & 1920$\times$1080 & 0.3m - 3m \\
\textit{Stereolabs ZED} & 60 & 1280$\times$720 & 1.5m - 20m \\
\textit{eCapture G53} & 30 & 640$\times$400 & 0.15m - 2m \\
\textit{DAVIS346 DVS} & - & 346$\times$260 & - \\
\hline
\end{tabular}
\label{tab:cameras}
\end{table}

The experiment consisted in placing each of the cameras in front of a scenario containing various objects and patterns. As the power consumption of the event cameras depends on the number of events triggered, we tested the \textit{DAVIS346 DVS} under three different conditions: \textit{Test1} involved moving the objects in the scene rapidly; \textit{Test2} involved moving the objects slowly, and \textit{Test3} involved keeping the objects stationary. To emulate the flapping strokes of the FWFR, all cameras were mounted on a benchmark that oscillated --varying the pitch angle in 60 degrees-- at different frequencies. To measure the pitch rate, we used a VectorNav VN-200 sensor, and for electrical consumption measurements, we used an INA219 wattmeter. This allowed us to analyze how different scenarios and flapping frequencies affect the camera's power consumption.

\begin{figure}[t]
\centering
\includegraphics[width=0.7\linewidth,page=1]{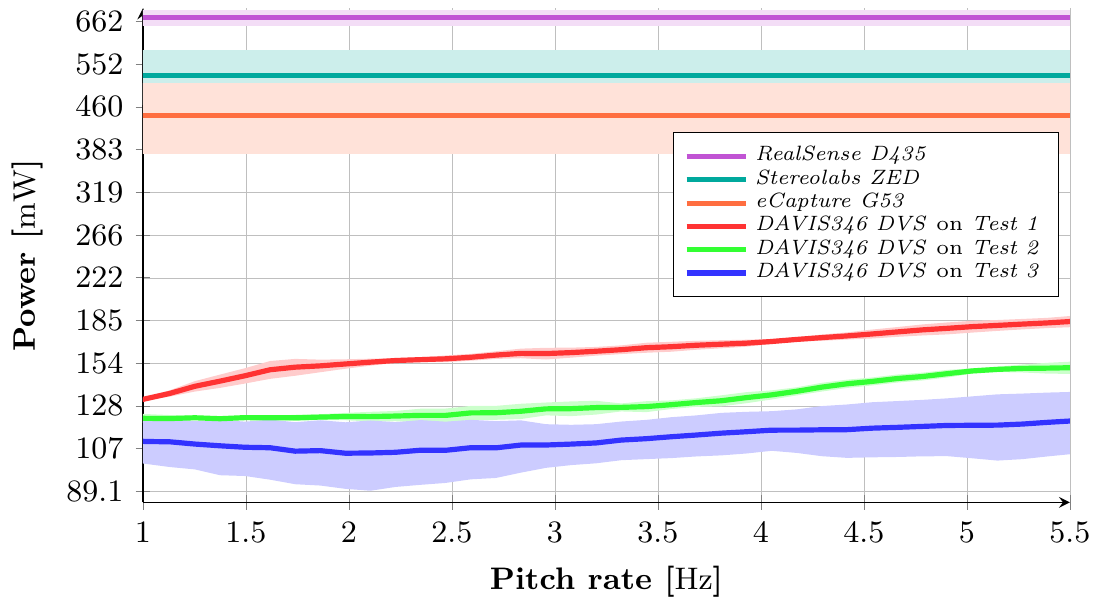}
\includegraphics[width=0.7\linewidth,page=2]{figures.pdf}
\vspace{-1em}
\caption{Energy consumption of the evaluated cameras as a function of pitch rates. The event rate is measured in millions of events per second (Meps).}
\label{fig:comparative}
\vspace{-1em}
\end{figure}

The experimental results, see Fig. \ref{fig:comparative}-top, clearly show that \textit{DAVIS346 DVS} offers significantly lower power consumption ($<$ 200 mW) compared to the rest of the cameras (in a range from 300 to 700 mW). The events generated in the experiments are shown in Fig. \ref{fig:comparative}-bottom. Although the energy consumed by the perception system is much lower than by other FWFR components (e.g., PMS motor), its analysis is relevant as it becomes significant in certain applications such as monitoring during long-distance gliding. In addition, it is interesting to consider that the dependency between power consumption and event generation rate is beneficial for FWFR applications since only certain parts of the scene (those with relevant information to process) trigger events and hence entail power consumption. In contrast, scenes with no relevant information (e.g., the open sky during long flights) trigger a very low number of events, involving negligible consumption.

\section{Conclusions}
\label{sec:conclusion}
This work investigated the power consumption of flight and monitoring components of a flapping-wing aerial robot. The power consumption was measured during the three different phases of the flight: \textit{launching}, \textit{ascending}, and \textit{cruising}. The consumed energy of the \textit{launching} phase and \textit{ascending} part is found 52.9\% and the \textit{cruising} flight 47.1\%. The vision monitoring section also analyzed event cameras and stereo cameras in terms of energy consumption. The experimental results confirmed event cameras as the most energy efficient perception sensor ($<$ 200 mW).

\section*{Acknowledgements}
This work was supported by the GRIFFIN ERC Advanced Grant 2017, Action 788247. Partial funding was obtained from the Spanish Projects ROBMIND (Ref. PDC2021-121524-I00), HAERA (PID2020-119027RB-I00), and the Plan Estatal de Investigación e Innovación of the Spanish Ministry (FPU19/04692).

\bibliographystyle{IEEEtran}
\bibliography{icra2023ws}

\end{document}